\documentclass[letterpaper, 10 pt, conference]{ieeeconf}  

\IEEEoverridecommandlockouts                              

\usepackage{amsmath} 
\usepackage{amssymb}  
\usepackage{graphicx}
\usepackage{algorithm}      
\usepackage{algpseudocode}  
\usepackage{xcolor}
\usepackage{tikz}
\usetikzlibrary{automata, positioning, arrows.meta}
\usepackage{subfig}
\usepackage{booktabs}
\usepackage{multirow}
\usepackage[flushmargin]{footmisc}

\title{\LARGE \bf
	DORA: Divergence-Oriented Data-Relay Algorithm \\
	for Partially Connected Robot Teams
}

\author{Jonathan Diller, Fernando Cladera, Camillo Jose Taylor, and Vijay Kumar
\thanks{*This work was supported by the ARL DCIST CRA W911NF-17-2-0181}
\thanks{All authors are with the GRASP Laboratory, University of Pennsylvania, Philadelphia, PA 19104, USA
        {\tt\small \{diller, fclad, cjtaylor, kumar\}@engineering.upenn.edu}}%
}

\begin{document}

\maketitle
\thispagestyle{empty}
\pagestyle{empty}

\begin{abstract}
	Teams of unmanned aerial vehicles (UAVs) deployed for search and monitoring missions frequently operate as partially connected networks, forcing each robot to trade off exploring the environment against relaying information to teammates. This tradeoff is especially acute when robots are semantically heterogeneous: an observation that appears uninformative to the robot that made it may be critical to a teammate with complementary detection capabilities. In this work, we formalize this setting as the heterogeneous mission-aware coverage (HMAC) problem, which couples complete multi-robot coverage of an area with capability-constrained mission-relevant target (MRT) discovery under intermittent communication. We then present DORA, a divergence-oriented data-relay algorithm that drives communication by the value of information to the team rather than by discovery alone. DORA quantifies the mission-relevant divergence between a robot's current information state and its estimate of each teammate's knowledge, capturing mission relevance, discovery novelty, sensor uncertainty, and the age of information. We evaluate DORA in simulation across four environments with differing object densities and spatial structure, and validate it on a physical UAV platform. Our results show that DORA improves MRT resolution delay by up to 74.8\% over traditional time-based communication scheduling methods.
\end{abstract}

\section{INTRODUCTION}
Teams of unmanned aerial vehicles (UAVs) are increasingly deployed to search large areas from the air, with applications that include search and rescue, security and intruder detection, environmental monitoring, and reconnaissance. By the nature of these tasks, the robot team usually spreads out over large deployment spaces in an effort to maximize coverage and parallelize monitoring. This often leads to partially connected teams and creates a tradeoff between searching the environment and relaying information.

This tradeoff is especially problematic when heterogeneity is semantic rather than geometric. For example, consider a team of UAVs searching an area for mission-relevant targets (MRTs). Every UAV can identify common object classes, such as cars, buildings, and roads, but each robot can evaluate only a subset of the MRT types that the team as a whole is tasked to discover. An observation itself may therefore have value to one teammate despite appearing uninformative to the robot that initially made the observation. 
A robot must determine not only whether it has discovered new information, but also whether that information is sufficiently valuable to a teammate to justify the cost of communicating. 

While the classical coverage path planning (CPP) problem is well understood~\cite{Choset2001Coverage, galceran2013survey}, these lines of work assume that observations follow from geometry: whatever a robot's sensor footprint passes over is fully observed. 
Conversely, existing communication-aware exploration methods either limit coverage by strictly requiring constant communication~\cite{Fink2012Robust, Tang2024Decentralized}, use precomputed rendezvous that ignore online discoveries~\cite{Hollinger2010Multi,Silva2024Communication}, or rely on mission-agnostic metrics such as time-lapse~\cite{Cladera2024Enabling,Miller2024Air}.

In this work, we formalize this setting as the heterogeneous mission-aware coverage (HMAC) problem: a team of UAVs with complementary MRT-detection capabilities must guarantee complete coverage of an area of interest and the discovery of every MRT present in it, while minimizing the mission completion time. 
HMAC strictly generalizes classical multi-robot coverage, reducing to the latter when all robots share the same detection capabilities, and couples it with capability-constrained inspection, which induces task-allocation and coordination subproblems that do not arise in conventional coverage.

To address this challenge, we introduce \textbf{DORA}: a divergence-oriented data-relay algorithm for partially connected robot teams based on the central idea that communication should be driven by the \emph{value of information to the team}, rather than by discovery alone. DORA explicitly quantifies the information divergence between a robot's current information state and its estimate of a teammate's knowledge, accounting for mission relevance, discovery novelty, sensor uncertainty, and the age of information. 
This paper makes the following contributions:
\begin{itemize}
	\item We introduce and formalize HMAC, a problem formulation that couples complete multi-robot coverage with capability-constrained MRT discovery (Sec.~\ref{sec:problem}).
	\item We develop DORA, a decentralized online planner for guiding the actions of partially connected, semantically heterogeneous teams (Sec.~\ref{sec:solution}).
	\item We evaluate our approach in simulation and field experiments, showing that DORA improves the balance between exploration and information relaying relative to mission-agnostic and time-based approaches for sharing information (Sec.~\ref{sec:eval}).
\end{itemize}

\section{RELATED WORK}

Classical coverage path planning (CPP) and its multi-robot extensions—which partition spaces to minimize makespan~\cite{Choset2001Coverage} or plan redundant sweeps to maximize detection~\cite{Hazon2008redundancy}—rely on a strictly geometric assumption: any region swept by a sensor is successfully observed~\cite{galceran2013survey}. This assumption breaks down in heterogeneous teams, where a robot's specific capabilities determine the value of an observation. Consequently, spatial coverage alone cannot guarantee that mission-relevant information is effectively gathered or delivered to the appropriate teammate.

Beyond strict area coverage, general multi-robot exploration and search methods, including frontier-based~\cite{Yamauchi1998Frontier}, coordinated~\cite{Burgard2005Coordinated}, and information-theoretic approaches~\cite{Charrow2015Information}, must constantly balance exploration progress against the need to share information. When addressing communication constraints, a conservative paradigm focuses on connectivity maintenance. These methods ensure continuous information flow via consensus-based control, graph-theoretic constraints, or joint communication-and-mobility control~\cite{Olfati2007Consensus, Zavlanos2011Graph, Fink2012Robust, Tang2024Decentralized}. However, tethering a team to a strict communication graph severely restricts motion, degrading coverage efficiency and precluding the wide-area deployments characteristic of search missions. 

To alleviate the constraints of persistent connectivity, intermittent-connectivity methods allow for temporary disconnections. One prominent approach is pre-planned rendezvous, where robots compute specific meeting locations and times to exchange information while otherwise exploring independently~\cite{Hollinger2010Multi, Silva2024Communication, Silva2025Intermittent}. The decision of when and where to communicate in these frameworks is primarily geometric or temporal, balancing travel costs against map synchronization requirements. While rendezvous planning effectively decouples communication from continuous line-of-sight, it typically relies on offline scheduling rather than the dynamic value of the information being gathered.

A more flexible sub-class of methods attempts to determine communication events online. Mechanisms for online synchronization vary widely, ranging from simple time-based triggers~\cite{Cladera2024Enabling, Miller2024Air} and purely opportunistic encounters~\cite{Racer2023Racer}, to occasionally backtracking to restore network links~\cite{Tellaroli2024Frontier, Gao2022Meeting}. Recent advances also explore learning to rendezvous during mapping tasks~\cite{Siang2024Implicit} or predicting teammate locations based on environmental structure and signal loss~\cite{Schack2024Sound}. Some communication-aware systems attempt to reduce message frequency by transmitting only when data provides sufficient novelty or general utility. Yet, these online decisions generally remain mission-agnostic; they treat information as uniformly valuable across the team, relying on arbitrary timeouts or raw observation counts rather than the semantic value of the accumulated data.

In heterogeneous systems, an observation redundant to one robot may be critical to another. DORA builds on decentralized opportunistic methods but replaces arbitrary synchronization triggers with a mission-aware formulation: a robot interrupts exploration only when the estimated information divergence between its state and a teammate's state exceeds a threshold. This ensures communication sacrifices efficiency only to materially update the team's knowledge.

\section{PROBLEM DEFINITION}\label{sec:problem}
We consider a team of $m$ UAVs, $\mathcal{R} = \{1,\ldots,m\}$, tasked with searching an area of interest $\mathcal{X} \subset \mathbb{R}^2$. The environment contains a set of objects $\mathcal{O} = \{o_1,\ldots,o_n\}$, where each object has a spatial position and an associated object class from a set of known classes $\mathcal{K}$ (e.g., vehicles, buildings, roads). A UAV may observe an object and determine its class, but additional capability may be required to evaluate whether that object constitutes an MRT.

Each UAV possesses a potentially distinct MRT-detection capability. Let $\mathcal{A}_j \subseteq \mathcal{K}$ denote the set of object classes whose mission-relevant status UAV $j$ is capable of evaluating. More generally, UAV $j$'s detection capability is represented as a mapping $g_j : \mathcal{O} \rightarrow \mathcal{Y}_j$, where $\mathcal{Y}_j$ describes the mission-relevant information that UAV $j$ can infer from an observation. Because capabilities across the team are heterogeneous, an observation may be sufficient for UAV $j$ to identify an MRT while remaining uninformative to UAV $i$.

Each UAV $j$ maintains an information state $I_j(t) = \{x_1,\ldots,x_{n_j}\}$, consisting of its collected observations. An observation $x_k$ includes the estimated object location, class, detection confidence, and timestamp. Observations are subject to sensor uncertainty, and objects may change position over time.

The team operates under intermittent connectivity. Two UAVs $i, j \in \mathcal{R}$ with communication ranges $r_i$ and $r_j$ can exchange data only when $\mathrm{dist}(i,j) \le \min(r_i, r_j)$. Let $\mathcal{C}(t) \subseteq \mathcal{R} \times \mathcal{R}$ denote the set of communicating UAV pairs at time $t$. When two UAVs communicate, they synchronize observations and routing state. Between communication events, UAV $j$ maintains an estimate $I_{ji}(t)$ of the information state currently possessed by UAV $i$. This estimate is initialized at the most recent synchronization event and degrades in accuracy as teammates make new discoveries.

Let $\mathcal{A}^* \subseteq \mathcal{O}$ denote the set of ground-truth MRTs present in $\mathcal{X}$. For any MRT $a \in \mathcal{A}^*$, we define two critical timestamps:
\begin{itemize}
	\item $t_a^{\mathrm{obs}}$: the earliest time at which any UAV in the team observes object $a$.
	\item $t_a^{\mathrm{id}}$: the earliest time at which a UAV $j$ capable of evaluating $a$ (i.e., $a$ matches capability $g_j$) receives the observation and correctly identifies it.
\end{itemize}
The quantity $\tau_a = t_a^{\mathrm{id}} - t_a^{\mathrm{obs}} \ge 0$ represents the information-induced relay delay.

Let $\mathcal{X}_j(t) \subseteq \mathcal{X}$ denote the cumulative spatial area swept by UAV $j$'s sensor footprint up to time $t$. The mission requires the team to achieve complete spatial coverage of $\mathcal{X}$ while minimizing the average time required to identify all MRTs.

Formally, we state the Heterogeneous Mission-Aware Coverage problem as:
\begin{align} 
	\min \quad & \frac{1}{\vert\mathcal{A}^*\vert} \sum_{a \in \mathcal{A}^*} \tau_a \label{eq:objective} \\
	\text{s.t.} \quad & \bigcup_{j \in \mathcal{R}} \mathcal{X}_j(T_m) = \mathcal{X}, \\ 
	& t_a^{\mathrm{id}} < \infty, \quad \forall a \in \mathcal{A}^*, \\ 
	& \text{kinematic and velocity constraints}\ \forall j \in \mathcal{R}, \\ 
	& \text{communication constraints defined by } \mathcal{C}(t), 
\end{align}
\noindent
where $T_m$ is the mission makespan at which complete spatial coverage is completed. Under this formulation, coverage is a strict requirement, and the objective explicitly rewards strategies that minimize the relay latency $\tau_a$ by propagating critical observations to capable teammates as early as possible.

\section{PROPOSED SOLUTION}\label{sec:solution}

\subsection{Information Gain}\label{subsec:info_gain}
We propose the following set of requirements for an information valuation model: (1) the model must map to a range that can be used for traceable decision making, (2) the model must be computationally efficient to evaluate, and (3) the model must quantify the novelty of information.

Suppose that robot \(j\) detects ground target \(i\) at time \(t_i\) and position \(\mathbf{x}_i \in \mathcal{X}\). Let the probability of an accurate detection be \(p_i\) and \(q_i = 1 - p_i\) be the probability that \(i\) was a false positive. Under the assumption that \(i\) was a true positive, let \(f_i(t)\) be the probability density function that describes the probability that \(i\) moves from position \(\mathbf{x}_i\) at time \(t\). The probability that \(i\) has moved between times \(t_i\) and \(t'\), where \(t' > t_i\), is defined by the cumulative distribution function of \(f_i(t)\), 
\begin{equation}
	F_i(t') = \int_{0}^{t' - t_i}f_i(t)\,dt,\label{eq:aoi}
\end{equation}
which is defined over elapsed time since detection ($t \geq 0$). The probability that target \(i\) is a true positive and is located at \(\mathbf{x}_i\) at time \(t'\) is
\begin{equation}
	p_e(i, t') = p_i\ \big(1-F_i(t')\big).
\end{equation}
We do not make assertions about \(i\) remaining within search space \(\mathcal{X}\) if \(i\) moves from \(\mathbf{x}_i\) and assume that $f_i(t)$ is known \textit{a priori}.

Let \(I = \{\mathbf{x}_1, \mathbf{x}_2,\ \dots\ ,\ \mathbf{x}_n\}\) be the set of all previously observed targets and \(\mathcal{I}\) be the Random Finite Set (RFS) of targets that currently exist in \(\mathcal{X}\). Random Finite Sets are sets containing a random number of possibly random vectors. We treat each \(i \in I\) as an independent binomial trial, where the probability that \(i\) is in \(\mathcal{I}\) at time \(t'\) is \(p_e(i, t')\) and the probability that \(i\) is not in \(\mathcal{I}\) is
\begin{equation}
	q_e(i, t') = 1 - p_e(i, t').
\end{equation}
Given that each binomial trial is independent, the probability that \(\tilde{n}\) trials are successful follows a Poisson binomial distribution (PBD), where \(0 \leq \tilde{n} \leq n\). Using the probability mass function (PMF) for the PBD introduced in~\cite{fernandez2010closed}, the probability that the cardinality of \(\mathcal{I}\) is \(\tilde{n}\) at time \(t'\), given the set of observations \(I\), is
\begin{align} \label{eq:P_r_fourier}
	&Pr\big[|\mathcal{I}| = \tilde{n}, t'\ \big|\ I\big] = \nonumber \\
	&\frac{1}{n \texttt{+} 1}\sum_{l = 0}^{n} \left\{ e^{\frac{\text{-2}\pi l \tilde{n} \sqrt{\text{-1}}}{n+1}} \prod_{i \in I} \left\{ p_e(i, t')e^{\frac{2\pi l \sqrt{\text{-1}}}{n+1}} + q_e(i, t')\right\} \right\}.
\end{align}
We define the RFS cardinality function as
\begin{equation}
	P(\tilde{n}, t', I) = 
	\begin{cases}
		Pr\big[|\mathcal{I}| = \tilde{n}, t'\ \big|\ I\big],	&0 \leq \tilde{n} \leq n\\
		0,	& n < \tilde{n}.\label{eq:p_ntI}
	\end{cases}
\end{equation}
\noindent
Observe that, by definition of the PMF of the PBD, 
\begin{equation}
	\sum_{\tilde{n}=0}^{n} P(\tilde{n}, t', I) = 1
\end{equation}
\noindent
and Equ.~\eqref{eq:p_ntI} is a valid probability measure.

Let \(I^{(k)}\subseteq I\) denote the subset of observations belonging to class \(k \in K\), where
\begin{align}
	I = \bigcup_{k \in K} I^{(k)}
\end{align}
and
\begin{align}
	I^{(a)} \cap I^{(b)} = \emptyset,
\end{align}
if $a \neq b$. Let \(\alpha_{ki} \in [0,1]\) denote the importance weight associated with class \(k\) for robot $i$, where
\begin{align}
	\sum_{k=1}^{K}\alpha_{ki} = 1,\quad \forall j \in \mathcal{R}.
\end{align}

For each class \(k\), we define the corresponding cardinality probability distribution
\begin{align}
	P^{(k)}(\tilde{n}, t', I^{(k)})
\end{align}
using Eq.~\eqref{eq:P_r_fourier} and Eq.~\eqref{eq:p_ntI} restricted to observations in \(I^{(k)}\).

Let \(I_j\) be the set of target observations currently held by robot \(j\) at time \(t'\). Let \(I_{ji}\subseteq I_j\) denote the subset of observations that robot \(j\) believes robot \(i\) possesses based on their most recent synchronization event. Equivalently, \(I_{ji}\) is robot \(j\)'s estimate of robot \(i\)'s information state. We define the information divergence between robot \(j\)'s current information state and its estimate of robot \(i\)'s information state for target class $k$ as the Hellinger distance between cardinality probabilities of the RFS generated by the observation sets
\begin{align}
	H_k(I_j,I_{ji})
		=
		\frac{1}{\sqrt{2}}
		\left\|
		\sqrt{P^{(k)}(\cdot,t',I_j^{(k)})}
		-
		\sqrt{P^{(k)}(\cdot,t',I_{ji}^{(k)})}
		\right\|_2.\label{eq:class_hellinger}
\end{align}
We define mission-relevant information divergence between \(j\)'s complete information state and its estimate of \(i\)'s complete information state as
\noindent
\begin{equation}
	H(I_j,I_{ji})
		=
		\sum_{k=1}^{K}
		\alpha_{ki}
		H_k(I_j,I_{ji}).\label{eq:hellinger}
\end{equation}
\noindent
To clarify, the divergence is not between independent world models. It is between a robot's current information state and the information state it believes a teammate possesses.

\noindent
\textbf{Information Valuation Requirements:} Since each class-specific Hellinger distance satisfies
\[
H_k(I_j,I_{ji})\in[0,1],
\]
and the class weights satisfy
\[
\alpha_{ki}\ge 0,
\qquad
\sum_{k=1}^{K}\alpha_{ki} = 1, \forall j \in \mathcal{R}
\]
the overall divergence \(H(I_j,I_{ji})\) is a convex combination of bounded distances and therefore also satisfies $H(I_j,I_{ji})\in[0,1]$, meeting requirement 1. Observe that Eq.~\eqref{eq:P_r_fourier} can be computed in \(\mathcal{O}(n^2)\) time, where \(n = |I_j|\), while \eqref{eq:p_ntI} tells us that the summation in \eqref{eq:hellinger} is zero for values above \(n\), which means that we can calculate \(H(I_j, I_{ji})\) in \(\mathcal{O}(n^3)\), meeting requirement 2. 
The Hellinger distance between two RFS \(I_j\) and \(I_{ji}\) initially grows quickly for making discoveries when there is little known about the environment but the delta in \(H(I_j, I_{ji})\) for making additional, unreported discoveries shrinks as the equation converges to 1; this naturally weights novel discoveries more heavily than incremental updates, satisfying requirement 3.

\noindent
\textbf{Triggering Communication:} 
We define the communication trigger for robot $j$ to communicate with robot $i$ as
\begin{equation}
	H(I_j, I_{ji}) \geq \varphi, \label{eq:trigger}
\end{equation}
where \(\varphi \in [0,1]\) is a communication threshold. Setting \(\varphi=0\) signifies that keeping teammates informed takes complete precedence over information gathering, whereas setting \(\varphi=1\) dictates that information gathering takes complete precedence over communication.

\begin{figure}[t]
    \centering
    \begin{tikzpicture}[
        node distance=1.4cm and 1.1cm,
        state/.style={circle, draw, thick, minimum size=1.0cm,
                      inner sep=1pt, font=\small, align=center},
        lbl/.style={font=\scriptsize, align=center},
        every edge/.style={draw, thick, -{Stealth[length=2.5mm]}}
    ]
        \node[state, initial, initial by arrow, initial where=left] (explore) {Explore};
        \node[state, above right=1.0cm and 1.9cm of explore] (relay) {Relay};
        \node[state, below right=0.2cm and 1.8cm of explore] (tour) {Neighbor\\Tour};
        \node[state, accepting, right=1.2cm of tour] (end) {End};

        \draw (explore) edge[bend left=20]
              node[lbl, pos=0.5, above left] {$\exists i: H(I_j, I_{ji}) \geq \varphi$} (relay);
        \draw (relay) edge[bend left=20]
              node[lbl, pos=0.9, yshift=4mm, below right] {$\nexists i:$\\$H(I_j, I_{ji}) \geq \varphi$} (explore);

        \draw (explore) edge[bend right=20, pos=0.9] node[lbl, yshift=-0.5mm, below left] {coverage\\complete} (tour);

        \draw (tour) edge[bend right=20]
              node[lbl, pos=0.4, right] {$\exists i:$ $H(I_j, I_{ji}) \geq \varphi$} (relay);

        \draw (tour) edge[above] node[lbl] {all robots\\complete} (end);
    \end{tikzpicture}
    \caption{Finite state machine depicting our decentralized information relaying algorithm.}
    \label{fig:state_machine}
    \vspace{-.5cm}
\end{figure}
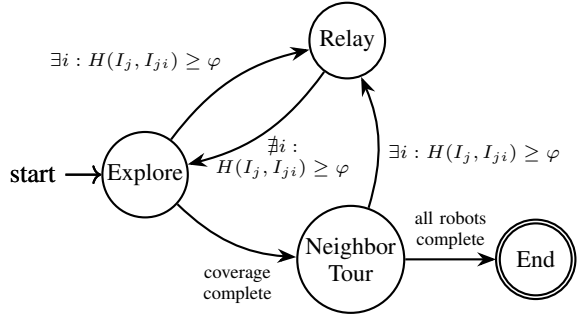

\subsection{Decentralized Information Relay Algorithm}\label{subsec:planner} 
Our algorithm, DORA, integrates our communication trigger mechanism into a decentralized algorithm for multi-robot coverage. We assume each robot follows a precomputed exploration route formulated as a Vehicle Routing Problem (VRP)~\cite{TothVehicle2014}. The algorithm determines exactly when a robot should divert from this baseline path to share mission-relevant discoveries with teammates.

\subsubsection{Opportunistic Communication}
Robot-to-robot communication is opportunistic, following a pattern similar to~\cite{Cladera2024Enabling}: transmissions occur only when two robots are within communication range of one another. This gossip-style mechanism allows information to diffuse through the team over multiple encounters without requiring any robot to maintain a centralized record.

In addition to observations, each robot shares its most recent routing schedule whenever it synchronizes with a peer. A routing schedule contains the robot's intended ad hoc meetings with other robots and its remaining exploration waypoints, together with the expected time at which the robot anticipates performing each of these tasks. Because schedules are exchanged during every synchronization event, each robot maintains a (possibly stale) estimate of its teammates' intended positions over time, which it uses when planning relay visits.

\subsubsection{Behavior Modes} 

Figure~\ref{fig:state_machine} illustrates the local state machine that dictates each robot's behavior. In short, the decentralized algorithm alternates between guiding the robot along its precomputed exploration path and temporarily diverting it into a relay mode whenever the communication trigger indicates a necessary information exchange with teammates.

\textbf{Explore:} The robots begin in the exploration state, following the
precomputed queue of exploration waypoints. While exploring, each robot $j$ iteratively evaluates the information divergence $H(I_j, I_{ji})$ between its
current information state and its estimate of every teammate $i$'s information
state. If $H(I_j, I_{ji}) \geq \varphi$ for some robot $i$, then $i$ becomes a
\textit{relay candidate} for $j$, and $j$ transitions to the relay state.

\textbf{Relay:} While in the relay state, robot $j$ must decide the order in
which to visit its relay candidates. When $j$ acquires more than one relay
candidate, it solves a traveling salesperson problem (TSP) over the candidates
using the most recently received routing schedule of each candidate $i$ to
estimate where and when each candidate can be intercepted. Robot $j$ then
visits each relay candidate in the order returned by the TSP solver. When
robot $j$ encounters a candidate $i$, the two robots synchronize their
databases using the opportunistic communication stack described above. After
synchronization, $H(I_j, I_{ji})$ drops below $\varphi$ and $i$ is removed
from the relay candidate list. Robot $j$ replans the TSP only if (1) a relay
candidate that is not the next scheduled visit is removed from the candidate
list, (2) a new robot $i$ with $H(I_j, I_{ji}) \geq \varphi$ is added, or (3) if there exists a robot $k$ that has advertised that it intends to attempt an ad hoc meeting with robot $i$ at an earlier time than $j$'s the intended encounter with $i$; in
all other cases the existing visiting order is retained, which avoids
unnecessary recomputation. Once all relay candidates have been visited and
removed (i.e., all databases are synchronized), robot $j$ returns to the
exploration state.

If robot $j$ arrives at an intercept location but candidate $i$ is not at the
position promised in its last shared schedule---for example, because $i$ diverted to perform its own relay visits---robot $j$ removes $i$ from its relay
candidate list and ignores the communication trigger for $i$ until it receives
an updated routing schedule from $i$. This prevents robot $j$ from
indefinitely chasing stale estimates of a teammate's position.

\textbf{Neighbor Tour:} When a robot exhausts its exploration waypoint queue, it transitions to the neighbor-tour phase. In this phase, the robot constructs a tour that visits each teammate that has not yet finished exploring and terminates at the base station. The purpose of this tour is to guarantee that information gathered during the final stages of exploration is delivered to the base station even if no relay candidate triggers fired. If the communication trigger fires while the robot is executing its neighbor tour, the robot returns to the relay state to service its relay candidates before resuming the tour. The mission ends when all robots have completed their exploration waypoints and returned to the base station.

\section{Evaluation} \label{sec:eval}
\begin{figure*}[ht!]
    \centering
    \subfloat[Camp]{\includegraphics[height=0.2\linewidth]{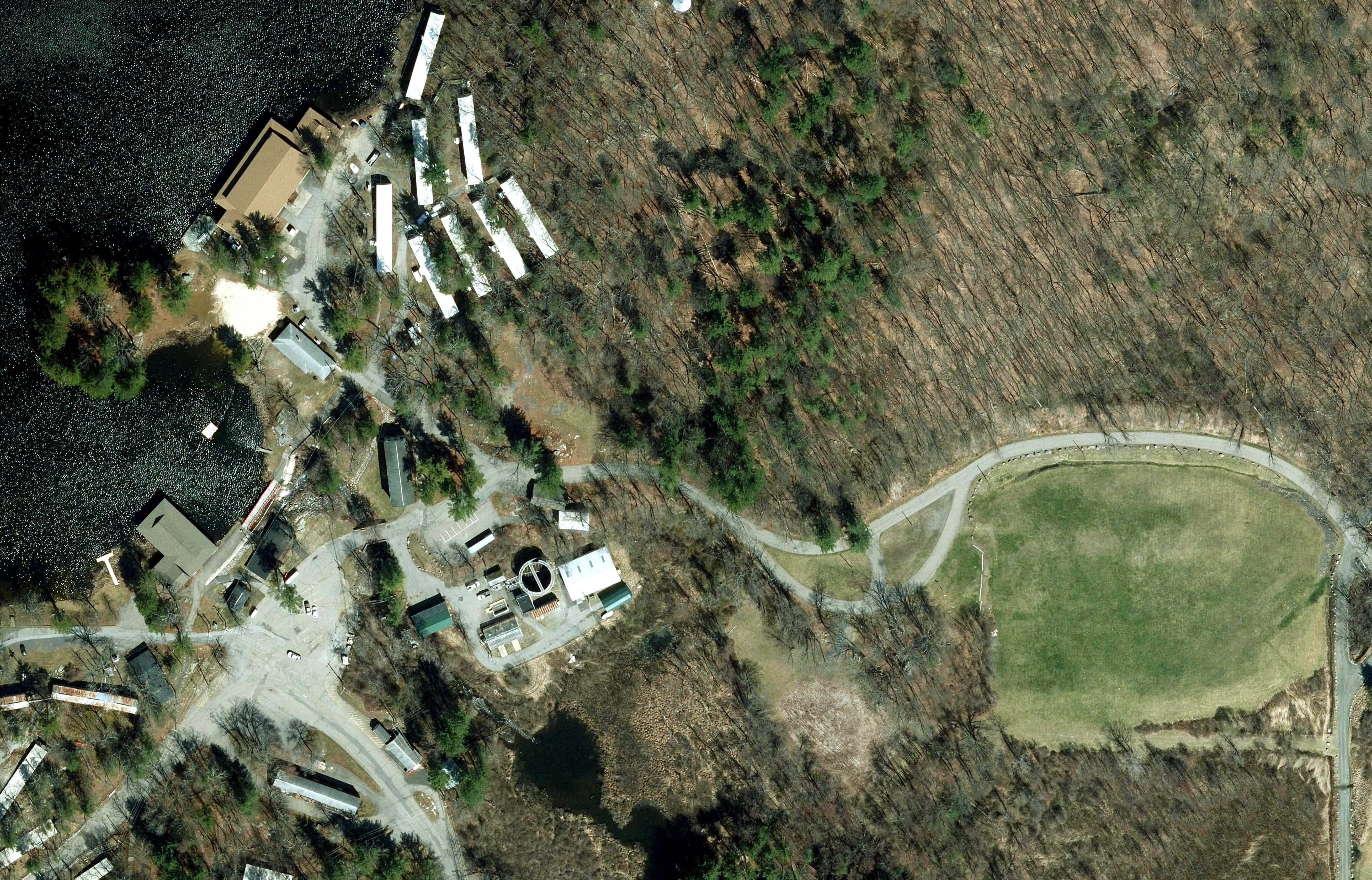}\label{fig:img1}}
    \,
    \subfloat[Suburbia]{\includegraphics[height=0.2\linewidth]{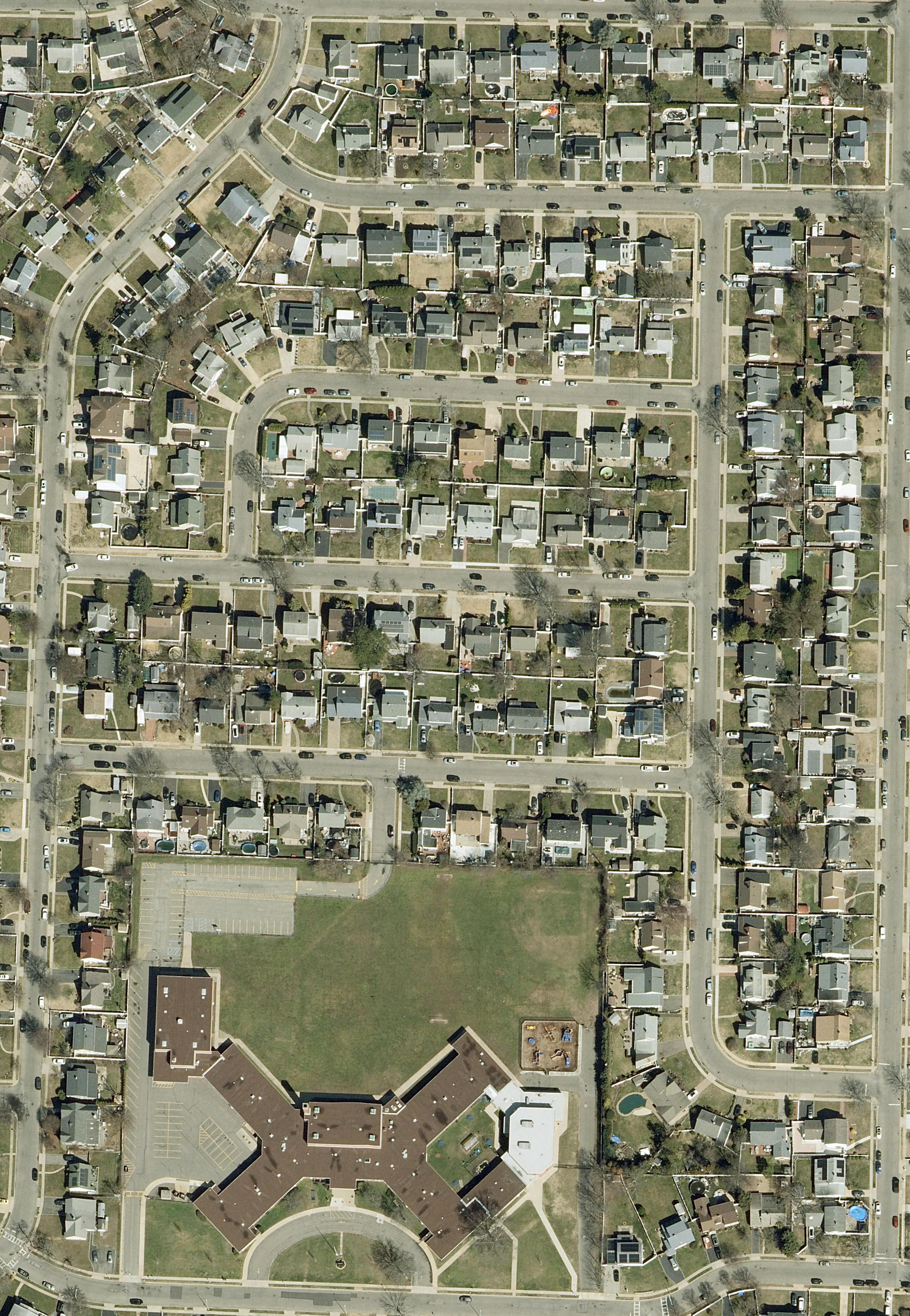}\label{fig:img2}}
    \,
    \subfloat[Farm]{\includegraphics[height=0.2\linewidth]{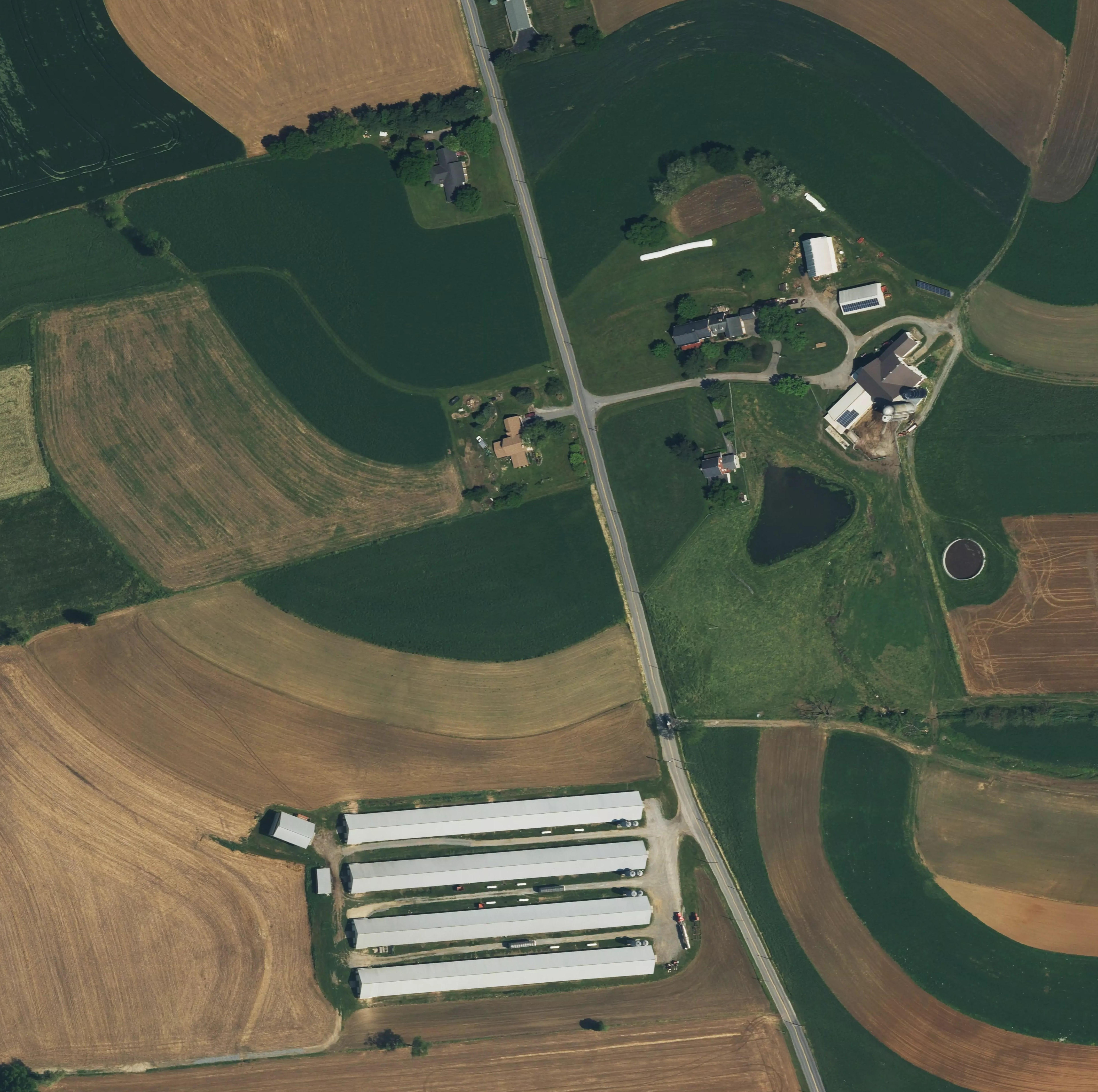}\label{fig:img3}}
    \,
    \subfloat[Business Park]{\includegraphics[height=0.2\linewidth]{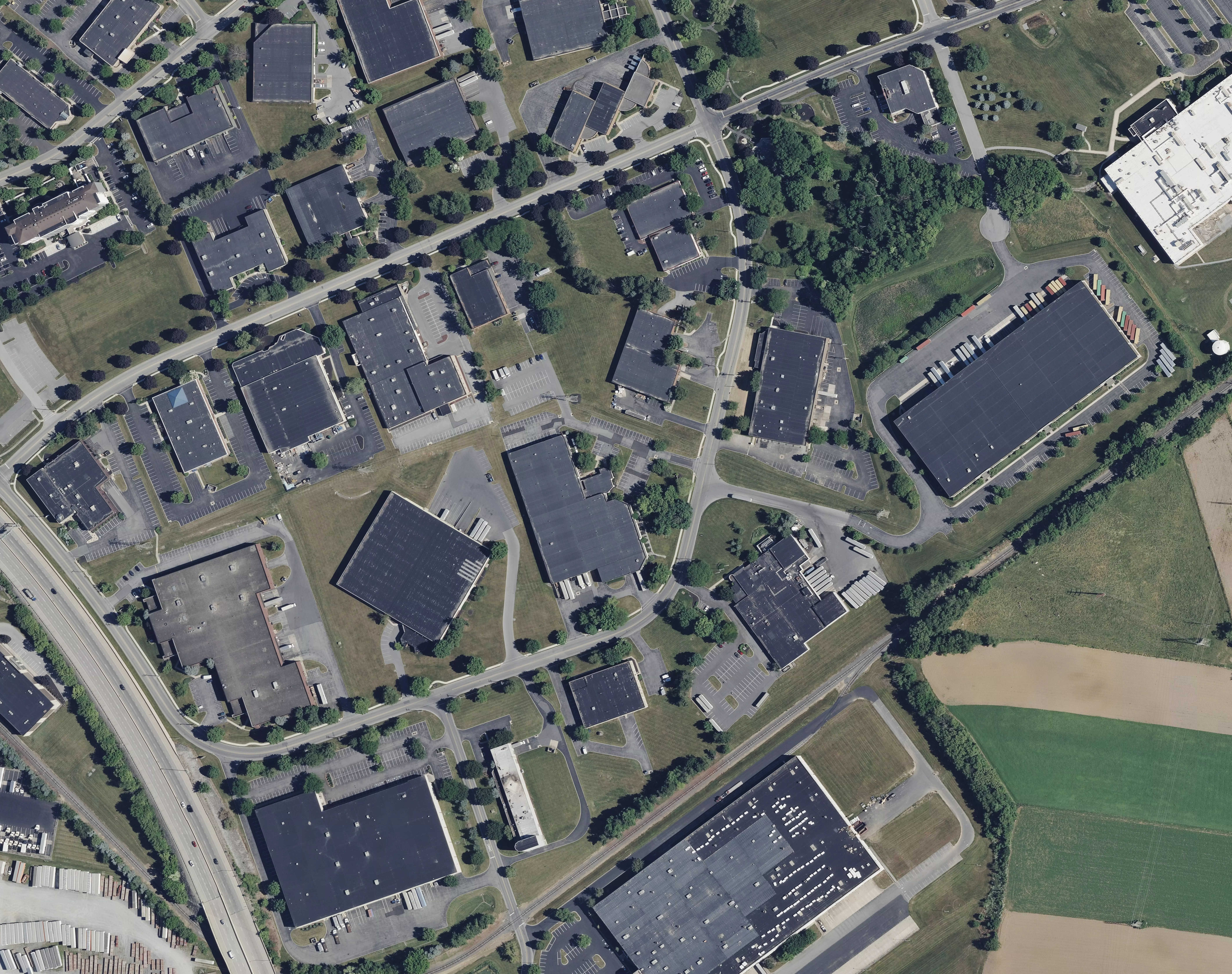}\label{fig:img4}}
    \caption{The four experimental areas used in simulation.}
    \label{fig:four_arenas}
\end{figure*}

We evaluate DORA in simulation and on physical robots, organized around three claims:
\begin{enumerate}
	\item[\textnormal{(E1)}] \emph{HMAC Performance and Scaling.} DORA outperforms baseline methods when applied to the HMAC problem and scales well with the size of the UAV team.   
	\item[\textnormal{(E2)}] \emph{Trigger Quality.} A divergence-based trigger achieves a better trade-off between mission time and team information freshness than time- and observation-count-based triggers across environments with very different object densities and spatial structure.
	\item[\textnormal{(E3)}] \emph{Feasibility.} The approach runs online, onboard physical robots, with results consistent with simulation.
\end{enumerate}

\subsection{Simulation Setup}
\label{sec:sim-setup}

\subsubsection{Environments}
\label{sec:environments}
We use four environments that vary greatly in layout, object distributions, and object densities. These four environments are: \emph{Camp}, a mixed terrain of forest, buildings, winding roads, and small parking lots; \emph{Suburbia}, dense with cars, houses, roads, and a school; \emph{Farm}, with sparse buildings, vehicles, and roads surrounded by mostly empty land; and \emph{Business Park}, with parking lots clustered around office buildings and connected by roads. Table~\ref{tab:environments} provides the statistics of each environment. The target movement probability density function, $f_i(t)$, follows a log-normal distribution parameterized by $\mu = 3.6 \times 10^3$ and $\sigma = 1.8 \times 10^3$ for cars, and $\mu = 10^9$ and $\sigma = 10^1$ for static buildings and road segments.

\begin{table}[t]
	\centering
	\caption{Object inventory of the four simulation environments.}
	\label{tab:environments}
	\begin{tabular}{lrrrr}
		\toprule
		Environment & Area $[\mathrm{km}^2]$ & Cars & Buildings & Roads \\
		\midrule
		Camp          & 0.12 & 11 & 31 & 55 \\
		Suburbia      & 0.17 & 364 & 187 & 131 \\
		Farm          & 0.24 & 10 & 22 & 64 \\
		Business Park & 0.37 & 78 & 20 & 189 \\
		\bottomrule
	\end{tabular}
\end{table}

\subsubsection{Robot Team, Capabilities, and Communication}
\label{sec:sim-team}
Unless otherwise stated, each simulation uses a team of $m = 2$ UAVs and a single base station. The base uses the same database and communication stack as the UAVs, as well as providing MRT detection capabilities, acting as a non-mobile teammate that still requires information sharing. Each UAV travels at a max speed of 10.0 m/s and has a sensor footprint of $256\pi\text{m}^2$. Communication follows the model of Sec.~\ref{sec:problem} with communication range $r_c = 75$ m for all UAVs and the base. The UAVs maintain a consistent flight level of 40 m above the ground while the base is located at ground level. Table~\ref{tab:alpha_weights} shows the peer-wise class priority weights, $\alpha_{ki}$, for each robot and the base.

\begin{table}[t]
	\centering
	\caption{Object class weights $\alpha_{ki}$ for each UAV and base.}
	\label{tab:alpha_weights}
	\setlength{\tabcolsep}{4.5pt}
	\begin{tabular}{l ccccccc}
		\toprule
		Class & Base & UAV$_1$ & UAV$_2$ & UAV$_3$ & UAV$_4$ & UAV$_5$ \\
		\midrule
		Cars			& 0.7 & 0.9 & 0.3 & 0.5 & 0.0 & 0.2 \\
		Road Segments	& 0.3 & 0.0 & 0.0 & 0.5 & 0.5 & 0.8 \\
		Buildings		& 0.0 & 0.1 & 0.7 & 0.0 & 0.5 & 0.0 \\
		\bottomrule
	\end{tabular}
\end{table}

\subsubsection{Baseline \& Ablations}
\label{sec:sim-baselines}
We consider three baselines: one based on a common information relaying approach found in the literature, and two ablations of DORA. All methods use DORA's decentralized tasking algorithm and differ from it only in how we trigger deviating from exploration to communicate with peers. The Time-Based (TB) baseline is based on works that use a time-based interaction mechanism~\cite{Cladera2024Enabling,Miller2024Air} and initiates an ad hoc meetings with teammate $i$ when the time elapsed since the last data synchronization event with $i$ exceeds a timeout $\varphi_{t} \geq 0$. 

The first ablation that we consider is the Mission-Agnostic (MA) approach, which assumes that all discoveries hold the same value uniformly across the entire team. The MA approach replaces $\alpha_{ki}$ in Equation~\ref{eq:hellinger} with $\frac{1}{|K|}$ to normalize the Hellinger distance across object classes while continuing to use Equation~\ref{eq:trigger} to trigger communication. Our second ablation is the Weighted-Count (WC) approach, which continues to use $\alpha_{ki}$ but removes the Hellinger distance calculation. Instead, WC calculates the class-wise difference in observations held by robot $j$ but not yet synchronized with robot $i$, triggering communication whenever
\begin{align}
	\sum_{k \in K} \alpha_{ki} |I_{j}^{(k)} \setminus I_{ji}^{(k)}| \geq \varphi_{o}
\end{align}
\noindent
for $\varphi_{o} \geq 0$.

\subsubsection{Metrics}
\label{sec:sim-metrics}
Our primary metric, as initially defined in Equation~\eqref{eq:objective}, is the average time required to identify all MRTs:
\begin{equation}
    \bar{\Delta}^{\mathrm{mrt}}
    = \frac{1}{|\mathcal{A}|}\sum_{o\in\mathcal{A}} \tau_o.
\end{equation}
\noindent
$\mathcal{A}$ is the set of all MRTs and $\tau_o$ is the time delta between $o$'s discovery and $o$ being delivered to a UAV capable of identifying that $o \in \mathcal{A}$.

The \emph{mission time}, $T_m$, is the time at which coverage is complete and all UAVs return to the base station. The \emph{mean information divergence}, $\bar{H}$, is the average $H(I_j, I_{ji})$ between teammate pairs (and the base) over the duration of the mission.

The \emph{$\alpha$-weighted delivery delay}, $\bar{\Delta}_{\alpha}$, measures how quickly mission-relevant information propagates through the team:
\begin{align}
	\bar{\Delta}_{\alpha} = \frac{1}{|\mathcal{R}||\mathcal{O}_{\mathrm{disc}}|} \sum_{j \in \mathcal{R}} \sum_{o \in \mathcal{O}_{\mathrm{disc}}}\alpha_{c(o)j}(t^{\mathrm{del}}_{oj} - t^{\mathrm{disc}}_o)
\end{align}
where $t^{\mathrm{disc}}_o$ is the time object $o$ is first observed by any UAV, $t^{\mathrm{del}}_{oj}$ is the time that $o$ is shared with robot $j$ for all robots in $\mathcal{R}$ other than the discoverer, and $c(o)$ is the class of $o$. Because $\alpha_{ki}$ encodes the peer-wise, class-level MRT prior, $\bar{\Delta}_{\alpha}$ can be interpreted as the expected delay before mission-relevant information reaches a teammate capable of processing it. We note that $\bar{\Delta}_{\alpha}$ considers all objects in $\mathcal{O}$ while $\bar{\Delta}^{\mathrm{mrt}}$ focuses on objects that qualify as MRTs.

\begin{figure*}[t]
    \centering
    \subfloat[Camp]{\includegraphics[width=0.2\textwidth]{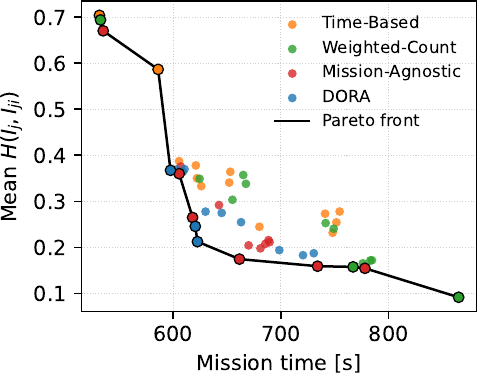}\label{fig:fnt_h_camp}}
    \hfil
    \subfloat[Suburbia]{\includegraphics[width=0.2\textwidth]{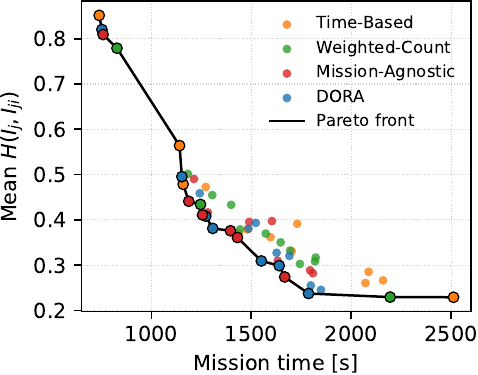}\label{fig:fnt_h_burb}}
    \hfil
    \subfloat[Farm]{\includegraphics[width=0.2\textwidth]{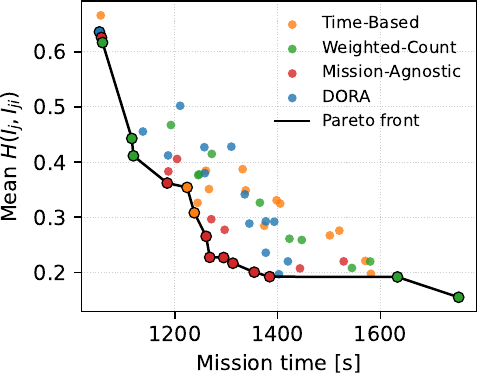}\label{fig:fnt_h_farm}}
    \hfil
    \subfloat[Business Park]{\includegraphics[width=0.2\textwidth]{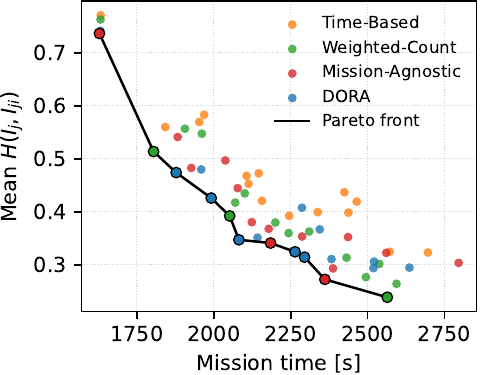}\label{fig:fnt_h_busi}}

    \subfloat[Camp]{\includegraphics[width=0.2\textwidth]{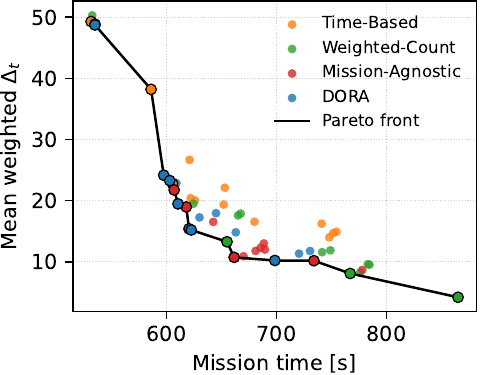}\label{fig:fnt_camp}}
    \hfil
    \subfloat[Suburbia]{\includegraphics[width=0.2\textwidth]{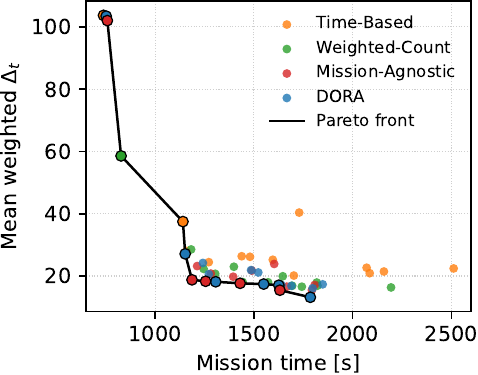}\label{fig:fnt_burb}}
    \hfil
    \subfloat[Farm]{\includegraphics[width=0.2\textwidth]{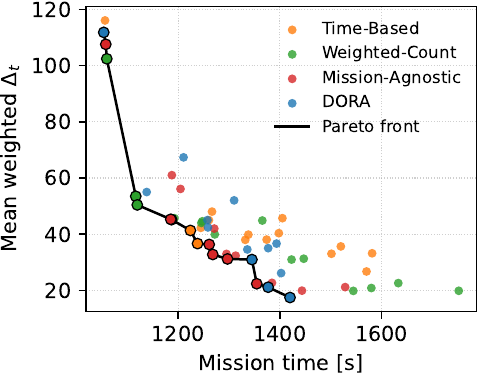}\label{fig:fnt_farm}}
    \hfil
    \subfloat[Business Park]{\includegraphics[width=0.2\textwidth]{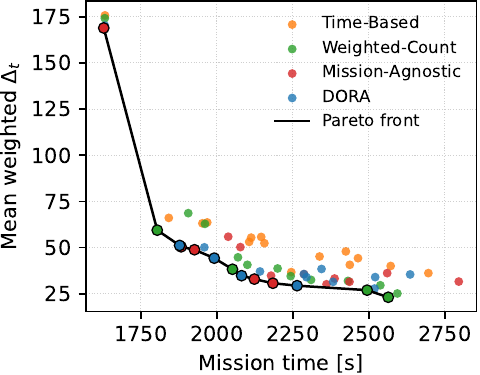}\label{fig:fnt_busi}}
    \caption{Pareto fronts between total mission time versus mean
    divergence $H(I_j, I_{ji})$ (top row), and mean weighted $\Delta_t$ (bottom row),
    for each arena.}
    \label{fig:pareto_front}
\end{figure*}

\subsection{Experiment 1: HMAC Performance and Scaling}
\label{sec:exp2}
In this section we evaluate the performance of each algorithm on the HMAC problem, where we want to minimize the time delay between discovering an object and identifying that it is an MRT. We randomly selected ``MRT'' objects from the set of all objects in the environment using the following method: (1)~draw an object class $k \in \mathcal{K}$ uniformly; (2)~draw an object $o \in \mathcal{O}_k$ uniformly; (3)~draw an identifier robot $j$ in proportion to the team's normalized-aggregate interest in that object class, excluding its discoverer. We then record the time delay between object discovery and reporting the observation to robot $j$. We repeat this process 100 times per condition and report the mean and standard deviation of the sampled delays in Table~\ref{tab:relay-delay}.

From the table, we see that the DORA algorithm consistently reduces the average relay delay compared to the baseline approaches in almost all environments except for the Farm. 
This shows us that the divergence metric helps in dense, class-heterogeneous environments where cardinality distributions are informative, while large and sparse setups may not be semantically rich enough for the metric to be as effective.

\begin{table}[t]
	\centering
	\caption{Simulation showing the time required to identify all MRTs, $\bar{\Delta}^{\mathrm{mrt}}$ (mean $\pm$ standard deviation), over 100 trials per cell.}
	\label{tab:relay-delay}
	\resizebox{\columnwidth}{!}{%
	\begin{tabular}{l cccc}
		\toprule
		& \multicolumn{4}{c}{Algorithm} \\
		\cmidrule(lr){2-5}
		Robots & DORA & MA & TB & WC \\
		\midrule
		\multicolumn{5}{c}{Camp} \\
		2 & \textbf{64.9} $\pm$ 68.1 & 73.0 $\pm$ 77.0 & 99.9 $\pm$ 71.5 & 117.8 $\pm$ 103.0 \\
		3 & \textbf{29.7} $\pm$ 37.6 & 35.3 $\pm$ 39.9 & 62.1 $\pm$ 60.7 & 43.6 $\pm$ 39.9 \\
		4 & \textbf{30.0} $\pm$ 40.0 & 34.7 $\pm$ 37.5 & 52.3 $\pm$ 60.2 & 36.8 $\pm$ 44.6 \\
		5 & \textbf{20.9} $\pm$ 25.2 & 33.4 $\pm$ 50.8 & 31.4 $\pm$ 41.6 & 25.7 $\pm$ 30.9 \\
		$\varphi$ & 0.72 & 0.88 & 182.0 & 6.0 \\
		\addlinespace
		\multicolumn{5}{c}{Suburbia} \\
		2 & \textbf{59.2} $\pm$ 49.8 & 69.5 $\pm$ 58.4 & 61.4 $\pm$ 43.6 & 81.5 $\pm$ 74.9 \\
		3 & \textbf{45.3} $\pm$ 44.6 & 59.6 $\pm$ 66.7 & 71.8 $\pm$ 57.6 & 51.3 $\pm$ 48.1 \\
		4 & \textbf{40.1} $\pm$ 41.1 & 65.6 $\pm$ 115.0 & 59.7 $\pm$ 47.8 & 46.8 $\pm$ 43.1 \\
		5 & \textbf{54.3} $\pm$ 69.1 & 58.5 $\pm$ 62.3 & 67.8 $\pm$ 51.6 & 80.8 $\pm$ 118.7 \\
		$\varphi$ & 0.92 & 0.88 & 110.0 & 21.0 \\
		\addlinespace
		\multicolumn{5}{c}{Farm} \\
		2 & 208.3 $\pm$ 185.4 & \textbf{147.7} $\pm$ 116.7 & 201.6 $\pm$ 97.9 & 191.7 $\pm$ 154.6 \\
		3 & 136.2 $\pm$ 97.3 & 127.4 $\pm$ 93.1 & 193.1 $\pm$ 109.7 & \textbf{119.6} $\pm$ 78.9 \\
		4 & 111.6 $\pm$ 106.6 & 170.1 $\pm$ 123.8 & 196.2 $\pm$ 120.5 & \textbf{87.0} $\pm$ 72.9 \\
		5 & 103.7 $\pm$ 104.8 & 106.8 $\pm$ 93.2 & 164.5 $\pm$ 107.9 & \textbf{86.6} $\pm$ 69.5 \\
		$\varphi$ & 0.84 & 0.84 & 410.0 & 5.0 \\
		\addlinespace
		\multicolumn{5}{c}{Business Park} \\
		2 & \textbf{184.1} $\pm$ 163.9 & 215.7 $\pm$ 157.0 & 731.1 $\pm$ 542.3 & 241.2 $\pm$ 186.4 \\
		3 & \textbf{101.2} $\pm$ 71.9 & 109.7 $\pm$ 93.2 & 147.2 $\pm$ 104.0 & 146.7 $\pm$ 98.4 \\
		4 & \textbf{137.5} $\pm$ 114.0 & 147.9 $\pm$ 134.7 & 194.4 $\pm$ 124.6 & 190.8 $\pm$ 132.1 \\
		5 & 106.9 $\pm$ 105.0 & 128.0 $\pm$ 155.3 & 126.8 $\pm$ 111.3 & \textbf{98.6} $\pm$ 78.7 \\
		$\varphi$ & 0.88 & 0.80 & 320.0 & 13.0 \\
		\bottomrule
	\end{tabular}%
	}
\end{table}

\subsection{Experiment 2: Time--Information-Freshness Trade-off}
\label{sec:exp1}
To demonstrate how DORA can balance exploration versus mission-aware information sharing, we mapped out the Pareto front between both  $\bar{\Delta}_{\alpha}$ and $\bar{H}$ versus the total time, $T_m$, for each environment. We varied the setting of $\varphi$ for DORA from 1 down to 0.5 and set the trigger range of the baseline methods to value ranges that give us a similar range for $T_m$. Observe that $\varphi = 1$ is equivalent to never deviating from exploration to communicate, giving us an implicit fourth baseline that is information-agnostic. 

Figure~\ref{fig:pareto_front} depicts our Pareto front results. Table~\ref{tab:front_stats} shows the percentage of points from each algorithm that make up the Pareto front (labeled as \textit{\% of front}) and percentage of runs for each algorithm that are within 5\% of the Pareto front (labeled as \textit{within 5\%}). DORA defines the front on two of the four environments in both $\bar{H}$ and $\Delta_t$, while its mission-agnostic ablation leads the front on the opposing two environments. While not always the defining solution of the Pareto front, DORA averages more runs within 5\% of the front than any of the baseline methods across three environments. The primary exception in performance is on the Farm, where DORA struggles with the large, sparse environment.

\begin{table}[t]
    \centering
    \setlength{\tabcolsep}{3.5pt}
    \caption{Pareto front statistics per algorithm and arena for both
    objective metrics: the percentage of the Pareto front belonging to each
    algorithm (front share), and the percentage of that algorithm's runs
    within 5\% of the front.}
    \label{tab:front_stats}
	\begin{tabular}{l l cccc}
		\toprule
			& & \multicolumn{4}{c}{Arena} \\
		\cmidrule(lr){3-6}
			& & Camp & Suburbia & Farm & Business Park \\
		\midrule
		& & \multicolumn{4}{c}{\itshape $\bar{H}$} \\
		\addlinespace[3pt]
		\multirow{2}{*}{TB}
			& {\scriptsize\itshape \% of front} $\uparrow$ & 14.3\% & 20.0\% & 12.5\% & 0.0\% \\
			& {\scriptsize\itshape within 5\%} $\uparrow$ & 23.1\% & 46.2\% & 20.0\% & 6.7\% \\
		\addlinespace[3pt]
		\multirow{2}{*}{WC}
			& & 21.4\% 			& 15.0\% 			& 31.2\% 			& 27.3\% \\
			& & 33.3\% 			& 46.2\% 			& 35.7\% 			& 26.7\% \\
		\addlinespace[2pt]
		\multirow{2}{*}{MA}
			& & \textbf{42.9\%} & 30.0\% 			& \textbf{50.0\%} 	& 27.3\% \\
			& & 53.8\% 			& 61.5\% 			& \textbf{64.3\%} 	& 42.9\% \\
		\addlinespace[2pt]
		\multirow{2}{*}{DORA}
			& & 21.4\% 			& \textbf{35.0\%} 	& 6.2\% 			& \textbf{45.5\%} \\
			& & \textbf{69.2\%} & \textbf{64.3\%} 	& 21.4\% 			& \textbf{57.1\%} \\
		\midrule
		& & \multicolumn{4}{c}{\itshape Mean weighted $\Delta_t$} \\
		\addlinespace[2pt]
		\multirow{2}{*}{TB}
			& {\scriptsize\itshape \% of front} $\uparrow$ & 16.7\% & 14.3\% & 12.5\% & 0.0\% \\
			& {\scriptsize\itshape within 5\%} $\uparrow$ & 38.5\% & 23.1\% & 26.7\% & 6.7\% \\
		\addlinespace[3pt]
		\multirow{2}{*}{WC}
			& & 16.7\% 			& 7.1\% 			& 18.8\% 			& 30.8\% \\
			& & 50.0\% 			& 23.1\% 			& 28.6\% 			& 40.0\% \\
		\addlinespace[2pt]
		\multirow{2}{*}{MA}
			& & 27.8\% 			& 35.7\% 			& \textbf{37.5\%}	& \textbf{38.5\%} \\
			& & 53.8\% 			& 38.5\% 			& \textbf{71.4\%} 	& 42.9\% \\
		\addlinespace[3pt]
		\multirow{2}{*}{DORA}
			& & \textbf{38.9\%} & \textbf{42.9\%} 	& 31.2\% 			& 30.8\% \\
			& & \textbf{61.5\%} & \textbf{50.0\%} 	& 50.0\% 			& \textbf{50.0\%} \\
		\bottomrule
	\end{tabular}
\end{table}

\subsection{Real-Robot Experiments}
\label{sec:real}

\begin{figure*}[t]
    \centering
    \subfloat[GPS trace of the UAV showing decision points]{\includegraphics[width=0.9\textwidth]{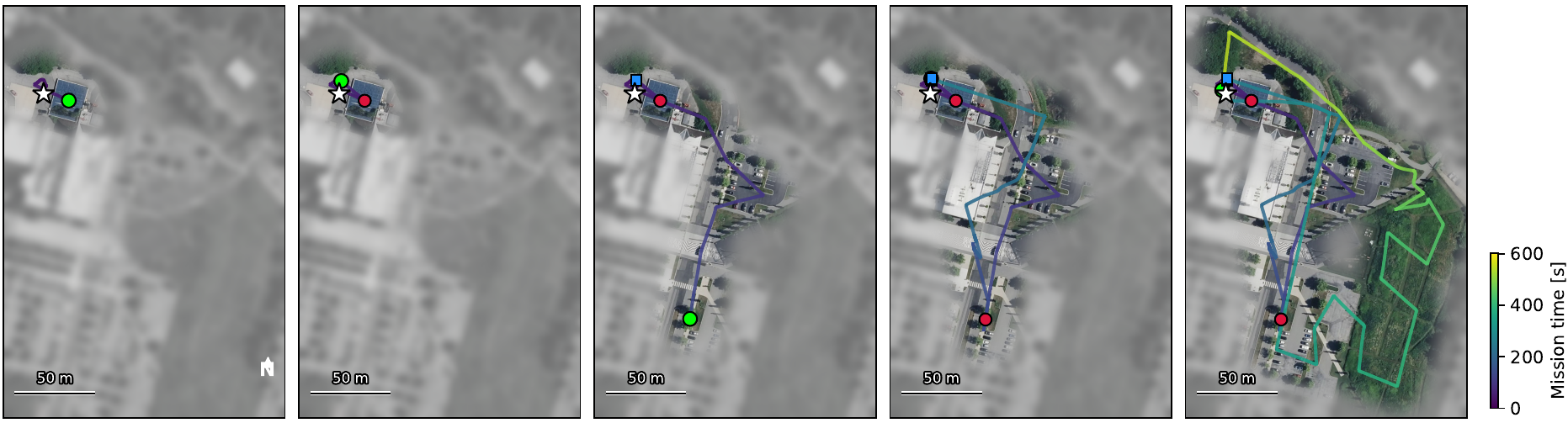}\label{fig:a}}

    \subfloat[Divergence (\(H(I_j,I_{ji})\)) between the UAV and the base station]{\includegraphics[width=.85\textwidth]{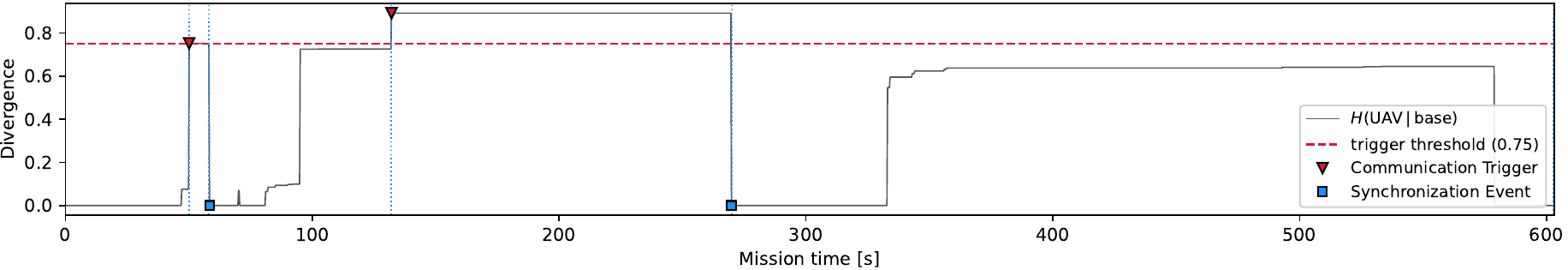}\label{fig:b}}
    \caption{GPS trace and decision points for our physical robot experiment. Red markers show the moment when communication is triggered (\(H(\text{UAV},\text{base}) \geq \varphi\)) and blue markers show when the UAV has synchronized its database with the base station. The emulated perception stack \emph{``discovers''} four cars at time marks: 50, $\sim$95, 132, and $\sim$330 seconds and communicates with the base three times at: 58, 270, and $\sim$575 seconds.}
    \label{fig:robot_experiment}
\end{figure*}

We validated DORA with a team of one UAV and one base station. The UAV is a custom platform powered by a PX4-based flight controller, and a NUC13 onboard computer. We used Rajant radios as the backbone for our communications system, relying on the radios API to provide information about the Received Signal Strength Indicator between two radios. We triggered communication between two nodes when 
$\text{RSSI} > 35$
and importantly, we only perform point-to-point communications.
For the purposes of this experiment, the perception stack was emulated in the robot. The UAV would \emph{``discover''} one object on the ground when flying within a pre-specified radius. All the other components of the system were running online on the robot. The robot and base class weights come from Table~\ref{tab:alpha_weights} for Base and UAV$_1$, and we set $\varphi = 0.75$.

Figure~\ref{fig:robot_experiment} depicts how the UAV explored the environment, and the divergence between the UAV's local database and information transmitted back to the base. The UAV flew autonomously along the depicted path except for three manual interventions due to safety concerns to keep the robot from flying over buildings and people. The UAV found five cars at times 0, 50, 95, 132, and 333 seconds. Two communication events were triggered at 50 and 132 seconds, aligning with discovering the second and third cars, respectively. The robot synchronized with the base at times 58 and 270 seconds, and at the very end of the mission at 579 seconds. 

The divergence graph demonstrates how discovering objects impacts DORA's decision-making process. The large jumps in divergence are from discovering cars and the small increments are from discovering road segments, reflecting the Base's preference for car observations over road segments. The graph shows how the novelty of discovering a car diminishes as the robot locates more cars, allowing the robot to better balance exploration and information sharing as its understanding of the environment changes online.

\section{CONCLUSIONS}
This paper presented \textbf{DORA}, a divergence-oriented data-relay algorithm that addresses the challenges posed by the HMAC problem. We show that DORA provides an effective solution to this problem through simulation experiments with teams of up to 5 UAVs in environments with differing object densities and spatial structure. We also performed field experiments to validate DORA in realistic settings with one UAV and one base station over real opportunistic communication links. DORA was able to effectively finish the coverage mission, returning to the base station when required to exchange information.

One of the biggest limitations of DORA is that it assumes obstacle-free motion. For instance, a relay operation estimates the position of other robots using the most recently received routing schedule. While this assumption holds for UAVs, it may not hold for Unmanned Ground Vehicles moving in an obstacle-rich space. We envision incorporating \emph{traversability} information estimation into our decision process. We leave this improvement for future work.

\addtolength{\textheight}{-12cm}   


\bibliographystyle{IEEEtran}
\bibliography{references}

\end{document}